\documentclass[letterpaper, 10 pt, conference]{ieeeconf}  %

\IEEEoverridecommandlockouts                              %

\usepackage{times}
\usepackage{booktabs}
\usepackage{xcolor}
\usepackage{algpseudocode}
\usepackage{algorithm}
\usepackage{subcaption}
\usepackage{makecell}
\usepackage{float}

\usepackage{amssymb}
\usepackage{graphicx}
\usepackage{dsfont}
\usepackage{lipsum}
\usepackage{amsmath}
\usepackage{multirow}
\usepackage{multicol}
\usepackage{boldline}

\let\labelindent\relax
\usepackage{enumitem}

\usepackage{siunitx}
\DeclareSIUnit\pixel{pixel}

\newcolumntype{P}[1]{>{\centering\arraybackslash}p{#1}}

\definecolor{LinkColor}{rgb}{0.12,0.49,0.85}
\makeatletter
\let\NAT@parse\undefined
\makeatother
\usepackage{hyperref}
\hypersetup{
    colorlinks=true,   
    linkcolor=black,
    citecolor=black,
    filecolor=black, 
    urlcolor=LinkColor,
    pdftitle={Semantic Segmentation Annotations for Aerial Field Robotics from Satellite Imagery},
}

\title{\LARGE \bf
Semantics from Space: Satellite-Guided Thermal Semantic Segmentation Annotation for Aerial Field Robots
}

\author{
Connor Lee, Saraswati Soedarmadji, Matthew Anderson, Anthony J. Clark, and 
Soon-Jo Chung%
\thanks{*This work was supported by the Office of Naval Research. The authors are with the California Institute of Technology. {Email: \texttt{\{clee, ssoedarm, matta, ajc, sjchung\}@caltech.edu}}. A.~J.~Clark is also with Pomona College.}
}

\begin{document}
\bstctlcite{IEEEexample:BSTcontrol}

\maketitle

\thispagestyle{empty}
\pagestyle{empty}

\begin{abstract}
We present a new method to automatically generate semantic segmentation annotations for thermal imagery captured from an aerial vehicle by utilizing satellite-derived data products alongside onboard global positioning and attitude estimates. This new capability overcomes the challenge of developing thermal semantic perception algorithms for field robots due to the lack of annotated thermal field datasets and the time and costs of manual annotation, enabling precise and rapid annotation of thermal data from field collection efforts at a massively-parallelizable scale. By incorporating a thermal-conditioned refinement step with visual foundation models, our approach can produce highly-precise semantic segmentation labels using low-resolution satellite land cover data for little-to-no cost. It achieves 98.5\% of the performance from using costly high-resolution options and demonstrates between 70-160\% improvement over popular zero-shot semantic segmentation methods based on large vision-language models currently used for generating annotations for RGB imagery. Code will be available at: \href{https://github.com/connorlee77/aerial-auto-segment}{https://github.com/connorlee77/aerial-auto-segment}.
\end{abstract}

\section{INTRODUCTION}

Uninhabited Aerial Vehicles (UAVs) have been extensively used in field robotic applications, including precision agriculture~\cite{pretto2020building}, wildlife conservation~\cite{bondi2020birdsai}, coastal mapping~\cite{brodie_bathy}, and wildfire management~\cite{jong2023wit}. To enable operations during nighttime and adverse weather conditions, UAVs can be equipped with long-wave thermal infrared cameras~\cite{delaune2019thermal, lee2023online} that provide dense scene perception in such settings. However, developing thermal scene perception for aerial field robotics requires ample data in order to train deep learning models for semantic segmentation~\cite{minaee2021image}. This poses a challenge due to the scarcity of in-domain thermal data capturing typical aerial field robotic operational areas such as deserts~\cite{8276269}, forests~\cite{jong2023wit}, and coastlines~\cite{nirgudkar2023massmind, brodie_bathy}. 

Although several thermal semantic segmentation datasets of urban scenes have been curated for autonomous driving applications~\cite{li2020segmenting, ha2017mfnet, vertens20bridging}, few datasets exist that specifically target natural environments from an aerial viewpoint~\cite{lee2024cart, lee2023online}. To compensate for limited thermal data, existing works leverage large, annotated RGB datasets via domain adaptation techniques like image translation~\cite{li2020segmenting} and domain confusion~\cite{gan2022unsupervised, msuda}, as well as online learning~\cite{lee2023online} for thermal test-time adaptation. Despite reducing reliance on thermal training data, such methods still require annotated thermal data for comprehensive evaluation and robustness testing. While thermal datasets exist for field environments, most lack annotations relevant for aerial semantic segmentation~\cite{jong2023wit, nirgudkar2023massmind, shivakumar2020pst900} besides~\cite{lee2024cart}. As a result, collecting and annotating thermal datasets for semantic segmentation is still necessary to further improve thermal scene perception results via supervised training.

Capturing and annotating thermal field data presents unique challenges. Unlike in RGB, publicly-available thermal imagery is scarce due to the high costs and specialized nature of thermal sensors. Consequently, relevant thermal imagery cannot be scraped from the web and field roboticists must travel to various locations for data collection. This process incurs significant time and financial expenses, as it requires extensive travel and permits for flying and data capture. Moreover, annotating thermal imagery adds further costs and delays due to its distinct visual characteristics. This requires multiple rounds of attentive expert review and re-annotation~\cite{lee2024cart}, and adds more time to the curation process.

In this study, we propose a method to significantly reduce the time and cost of annotating aerial thermal field imagery for semantic segmentation. \textit{Main contributions:}
\textbf{1.} An algorithm that automatically generates high-quality segmentation labels for aerial thermal imagery using estimated camera pose and satellite-derived data.
\textbf{2.} Experiments comparing segmentation labels generated from various satellite-derived data products, demonstrating competitive results with free options.
\textbf{3.} Extensive ablation studies showcasing the robustness of our method to noisy camera pose estimation and temporal misalignments between thermal and satellite imagery.
\textbf{4.} A demonstration for aerial field robot perception by training a semantic segmentation network solely on labels generated using our method, yielding promising results.

\begin{figure*}[ht]
    \centering
    \includegraphics[width=\linewidth]{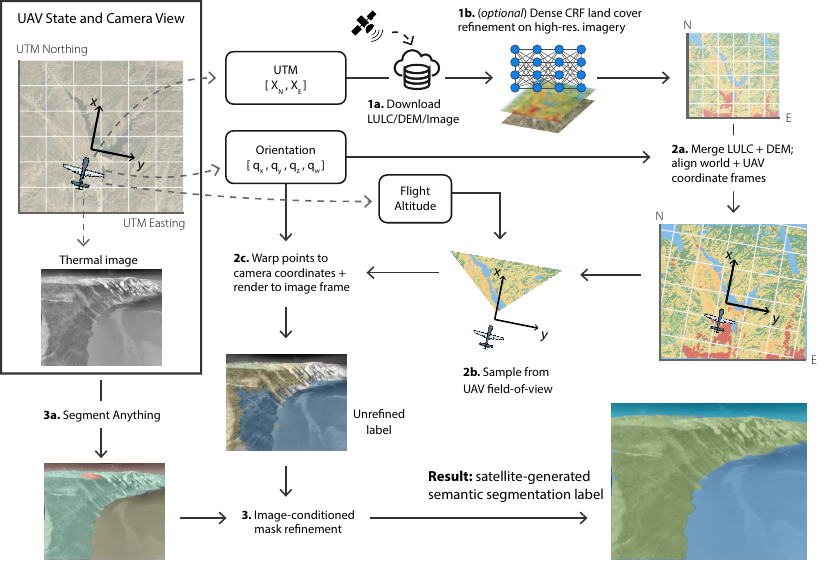}
    \caption{Proposed pipeline for automatically generating semantic segmentation annotations from satellite-derived data. Coarse segmentation labels for thermal images are rendered from Land Use and Land Cover (LULC) datasets and Digital Elevation Maps (DEM). The labels are refined using Segment Anything~\cite{kirillov2023segany} to capture fine details between segmentation instances.}
    \label{fig:flowchart}
\end{figure*}
\section{Related Work}
\label{sec:related-work}

\textbf{Semantic Segmentation:}
Semantic segmentation models perform per-pixel classification and are typically built upon convolutional neural networks and transformer architectures~\cite{chen2017rethinking, he2017mask, liu2021swin}. While conventional fully-supervised models achieve impressive results, they need large annotated training datasets for generalization. In applications like thermal semantic segmentation where labeled data is scarce, unsupervised domain adaptation (UDA) techniques are often employed. UDA methods like~\cite{li2020segmenting} synthesize labeled thermal training data from existing RGB datasets via image translation, while other works~\cite{gan2022unsupervised, msuda} leverage RGB training for thermal inference by maximizing RGB-thermal domain confusion during training. However, UDA methods still face challenges: they require significant target domain data and thermal annotations for evaluation, and typically exhibit lower performance compared to fully-supervised methods~\cite{mehra2021understanding}. 

Alternatively, recent large vision-language models like ODISE~\cite{xu2023open} and OV-Seg~\cite{liang2023open} can perform zero-shot semantic segmentation across the RGB spectrum by leveraging user-provided text prompts. Similarly, the Segment Anything Model~\cite{kirillov2023segany} (SAM) can provide precise segmentations for any object but lacks semantic information. In general, the zero-shot semantic segmentation methods perform worse on thermal imagery compared to RGB~\cite{lee2024cart}. Despite this, \cite{lee2024cart} finds that SAM can perform well in a semantic segmentation task if its segmentation outputs are assigned ground truth class labels. We leverage this finding in our approach. 

\textbf{Automatic Semantic Segmentation Annotation:}
Most works using automatic semantic segmentation labeling can be found in self-training and self-supervised learning literature. However, many focus on specialized applications with niche classes~\cite{lee2023online, daftry2018online} and are not relevant for general scene segmentation. For generalized semantic segmentation tasks, \cite{araslanov2021self} self-trains their model using noisy labels predicted by their network for intra-RGB domain adaptation. In contrast, \cite{yu2023weakly} adopts an incremental training approach and utilizes humans to select good network outputs as annotations and manually correct bad ones before retraining. Other works manually annotate a subset of frames in video data, before propagating them to remaining frames using optical flow~\cite{marcu2020semantics} or learned generative models~\cite{berg2019annotate}.

As discussed, visual foundation models can also be used for annotation efforts. Zero-shot semantic segmentation models~\cite{xu2023open, liang2023open, scaleafm1} are being used to provide labels, but do not transfer directly to non-RGB domains. \cite{gallagher2024multispectral} generates object detection labels for thermal imagery by using SAM on aligned RGB images and does not work in low-light settings.

In contrast to other works, \cite{braun2019combining} uses 3D information to generate semantic segmentation labels for construction sites and is most similar to our work. They register Building Information Models with point clouds from photogrammetry and render the labeled 3D points to an image frame. Unlike other works, methods like this operate independently of an image and can work for any imaging modality.

\section{Preliminaries}
In this section, we briefly go over the different satellite-derived data products we use in our approach (Sec.~\ref{sec:approach}).

\textbf{Land Use and Land Cover Datasets:}
Publicly-available Land Use and Land Cover (LULC) datasets like Dynamic World~\cite{brown2022dynamic} and Impact Observatory~\cite{9553499} derive from satellite rasters obtained through the Sentinel-2 program. These datasets have a low spatial resolution of \qty{10}{\meter\per\pixel} but have global coverage, and are updated using semantic segmentation networks that use multiple data bands for landcover classification. While daily coverage is possible, availability depends on factors like cloud coverage. 

In contrast, high-resolution LULC like the Chesapeake Bay Program~\cite{robinson2019large} and OpenEarthMap~\cite{xia2023openearthmap} offer sub-meter resolution but are limited in geographical and temporal coverage. While segmentation models can be trained on these datasets with high-resolution imagery, they may not generalize to different geographical areas. 

\textbf{High-Resolution Raster Imagery:}
These include imagery from aerial vehicles and satellites. Aerial imagery providers include the National Agricultural Imagery Program (NAIP)~\cite{NAIP} while satellite imagery comes from providers like Planet, Maxar, and Airbus. Image resolutions range from \qtyrange{0.3}{3}{\meter\per\pixel}. Imagery can be available daily at a premium cost while free alternatives are captured triannually.

\textbf{Lidar-Derived 3D Data Products:}
Digital surface (DSM) and digital elevation models (DEM) are raster data whose values denote the height at the corresponding geographic location. DSMs consider features above the ground like foliage and rocky terrain while DEMs report bare earth elevation. In this work, we use DEMs and DSMs with \qtyrange{1}{3}{\meter\per\pixel} resolution from the 3D Elevation Program (3DEP) from the United States Geological Survey~\cite{3dep} .

\section{Approach}
\label{sec:approach}
We present a 3 step method to automatically generate semantic segmentation annotations for thermal images captured from an aerial vehicle using satellite-derived data (Fig.~\ref{fig:flowchart}). 

\subsection{Step 1: Generating 3D Semantic Maps from Satellite Data}
We start by downloading relevant satellite data (LULC rasters, DEM or DSM, and high-resolution imagery) around the aerial vehicle's global position and resample them to the highest resolution via bicubic interpolation. To simplify future calculations, we convert to UTM coordinates before merging the DEM and LULC rasters.
Since current freely-available LULC data is low resolution (\qty{10}{\meter}), we optionally refine them by conditioning on high resolution imagery as described below. Alternatively, high-resolution LULC can also be created using a pretrained LULC segmentation network on high resolution imagery (see Sec.~\ref{sec:lulc-network-gen}).

\textbf{Land-Use-Land-Cover Refinement:}
\begin{figure*}[htbp]
    \centering
    \includegraphics[width=\linewidth]{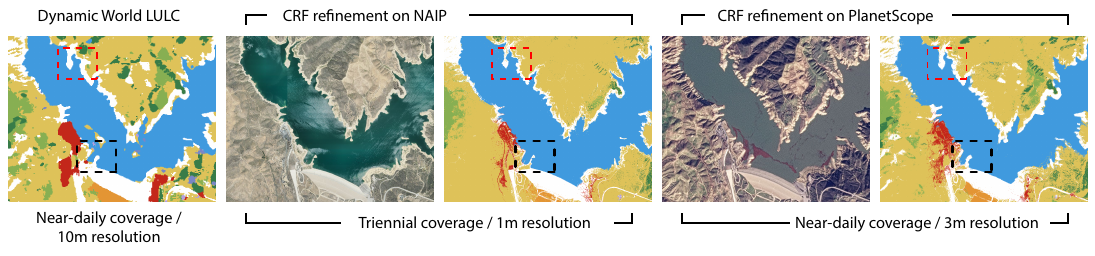}
    \caption{Dense CRF refinement of Dynamic World land cover raster using NAIP and PlanetScope imagery of Castaic Lake, CA. Results via PlanetScope convey the actual scenery at time of thermal image capture due to its high revisit frequency but at a lower \qty{3}{m} spatial resolution. NAIP refinement offers \qty{1}{m} resolution but is susceptible to changes in the terrain (notably, water levels of lakes) due to its triennial capture cycle. Zoom in to see key differences (outlined in dashed boxes).}
    \label{fig:lulc-refinement}
\end{figure*}
We use dense conditional random fields~\cite{krahenbuhl2011efficient} (CRF) to refine \qty{10}{m} resolution LULC rasters with \qty{1}{m}-\qty{3}{m} resolution aerial imagery (Fig.~\ref{fig:lulc-refinement}). To summarize, a dense CRF is defined by a Boltzmann distribution with energy function
\begin{equation}
    E(\mathbf{X|I}) = \sum_i \psi_u(x_i|I_i) + \sum_{i<j} \psi_p(x_i, x_j|I_i) 
\end{equation}
This function models the relationship between labels $\mathbf{x \in X}$ and the conditioning image $I \in \mathbb{R}^{H\times W \times C}$. Here, $\psi_u$ is a unary potential taken to be raw logits from a semantic segmentation network and $\psi_p$ is a pairwise potential that encourages label consistency among adjacent pixels with similar intensities. 

In our method, we set $\psi_u$ to be the logits from the model that generated our LULC labels. Like~\cite{kamnitsas2017efficient}, we use a generalized $\psi_p$ to condition on multi-band raster images:
\begin{multline}
    \hspace{-.3cm}\psi_p(\mathbf{f_i, f_j}) =  \mu\cdot\Biggl[ 
        w^{(1)} 
                \exp{\biggl(-\frac{1}{2}\mathbf{\bar{p}}_{ij}^\top \mathbf{\Sigma_\alpha} \mathbf{\bar{p}}_{ij} 
                - \frac{1}{2}\mathbf{\bar{I}}_{ij}^\top \mathbf{\Sigma_\beta} \mathbf{\bar{I}}_{ij} \biggr)} \\
        + w^{(2)} 
                \exp{\biggl(-\frac{1}{2}\mathbf{\bar{p}}_{ij}^\top \mathbf{\Sigma_\gamma} \mathbf{\bar{p}}_{ij} \biggr)} \Biggr]
\end{multline}
Here, $\mathbf{\bar{p}}_{ij} = [\mathbf{p}_i - \mathbf{p}_j] \in \mathbb{R}^3$ is the difference between positions of pixel $i$ and $j$, and $\mathbf{\bar{I}}_{ij} = [\mathbf{I}_i - \mathbf{I}_j] \in \mathbb{R}^C$ is the difference between image features at pixels $i$ and $j$. We set $\mu$ as the standard Potts compatibility function~\cite{krahenbuhl2011efficient}.

We optimize the CRF by tuning weight parameters $w^{(1)}$ and $w^{(2)}$, and the Gaussian bandwidth parameters ${\mathbf{\Sigma_\alpha} = \theta_\alpha}$, ${\mathbf{\Sigma_\gamma} = \theta_\gamma}$, and ${\mathbf{\Sigma_\beta} = \textrm{diag}\bigl(\theta_\beta^{(1)}, ..., \theta_\beta^{(C)}\bigr)}$. We minimize the boundary loss~\cite{bokhovkin2019boundary} and 
use this instead of cross entropy to account for imprecise labels at class boundaries due to low LULC resolution.

\subsection{Step 2: LULC Projection to Aerial Camera Image Frame}
To generate an LULC-derived semantic label for an image at time $t$, we start by transforming the world coordinates of each pixel $\mathbf{X}_w\in \mathbb{R}^3$ into the camera coordinate frame. This requires the position of the host vehicle $\mathbf{x}_t^{\mathrm{uav}}\in \mathbb{R}^3$, taken from the onboard EKF-fused GPS position and barometric altitude, the orientation quaternion $\mathbf{q}_t \in \mathbb{H}$, taken from the EKF-fused IMU readings, and the offset between the aircraft and camera reference points. Using the calibrated camera intrinsic matrix $\mathbf{K}$, we can then project to image coordinates $\mathbf{x}_t^c \in \mathbb{Z}^2$. Formally, this is
\begin{equation}
    \begin{bmatrix}
        \mathbf{x}_t^c \\
        1 
    \end{bmatrix} 
    =
    \mathbf{K}
    \begin{bmatrix}
        \mathbf{R}(\mathbf{q}_t)  & \mathbf{T}( \mathbf{x}_t^{\mathrm{uav}} )  \\
        \mathbf{0}_{1 \times 3} & 1 \\
    \end{bmatrix}
    \begin{bmatrix}
        \mathbf{X}_w \\
        1
    \end{bmatrix}
\end{equation}
where $\mathbf{R}$ is a rotation matrix and $\mathbf{T}$ is a translation vector. We use OpenGL to render the projected LULC, using 3D coordinates as vertices, associated class labels as vertex colors, and depth-testing to avoid rendering occluded semantics.

To optimize memory and speed, we only consider 3D semantics within a specified distance in front of and on both sides of the camera. We also exponentially increase spacing between sampled vertices as distance from the camera increases, exploiting the compression of far-field points in the image frame. This enables us to use only $250\times200$ points when rendering within a \qty{10}{km}$\times$\qty{8}{km} bounding box.

\begin{algorithm}[ht]
    \caption{SAM-based Label Refinement}\label{alg:sam-refinement}
    {\fontsize{9}{11}\selectfont
    \begin{algorithmic}[1]
        \State \textbf{Input:} Projected (unrefined) label mask $L \in \mathbb{N}^{H\times W}$, \\ \qquad\quad Thermal image $I \in \mathbb{R}^{H\times W}$
        \State \textbf{Output:} Refined semantic segmentation label $M$
        \State \textbf{Initialize:} Segment Anything Model $f_{\mathrm{sam}}$
        \\
        \State $\bigl\{M_{\mathrm{sam}}^i\bigr\}_{0}^N \gets f_{\mathrm{sam}}(I)$ \Comment{SAM produces binary masks}
        \State Initialize zero-array $M$ of size $H \times W$ 
        \For{$m_{\mathrm{sam}} \in \{M_{\mathrm{sam}}^i\}_{0}^N$}
            \State $x_{idx} \gets [m_{\mathrm{sam}} == 1]$  \Comment{Get mask indices}
            \State $y_{\mathrm{cls}} \gets L[x_{idx}].\mathrm{mode()}$ \Comment{Find most freq. class in mask}
            \State $M[x_{idx}] = y_{\mathrm{cls}}$
        \EndFor
        \State \textbf{return} $M$
    \end{algorithmic}
    }
\end{algorithm}

\subsection{Step 3: Rendered Label Refinement}
\label{sec:renderedlabelrefinement}
Though the semantic segmentation labels have been rendered, they do not align well with the thermal images. This is primarily caused by poor spatial resolution and temporal misalignment, but could also stem from errors in LULC label generation and camera pose estimation. To improve alignment, we refine the labels by generating binary segmentation masks of the corresponding thermal image using the Segment Anything Model. Then, we assign each mask a semantic class based on the most prevalent LULC class within it (Alg.~\ref{alg:sam-refinement}). 

\begin{figure*}[htbp]
    \centering
    \includegraphics[width=\linewidth]{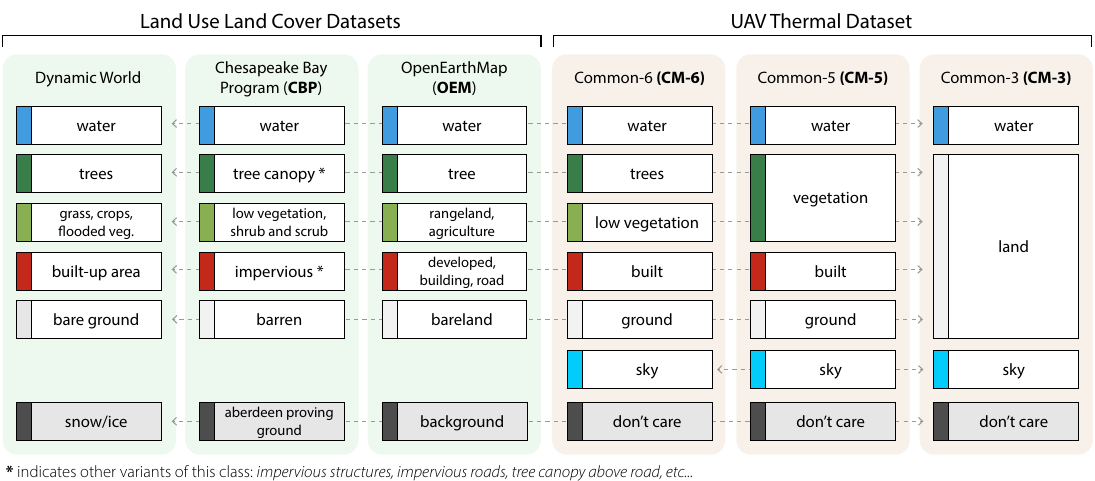}
    \caption{Class mappings between LULC datasets and our ground truth evaluation set. The UAV thermal dataset is from~\cite{lee2024cart}.}\label{fig:class-mappings}
\end{figure*}

\section{Low Altitude Aerial Dataset}
\label{sec:dataset}
We test our method using a thermal field robotics dataset, which includes off-nadir (20\textdegree-45\textdegree) aerial views of rivers (Kentucky River, KY and Colorado River, CA), lakes (Castaic Lake, CA), and coastal (Duck, NC) areas across the United States~\cite{lee2023online, lee2024cart}. The dataset, captured from a multirotor, comprises 15 flight trajectories ranging from \qty{40}{m} to \qty{100}{m} in altitude and contains time-synchronized thermal imagery, GPS, and IMU measurements. Four trajectories are excluded from testing due to GPS data collection errors. While the dataset provides ground truth semantic segmentation annotations for 10 classes, we condense the classes into 6 categories in order to better conform with land cover classes. We end up with ground truth, 6-class semantic segmentation labels for 1304 sub-sampled images (CM-6) and further condense the classes again to create two additional class-sets, CM-5 (5 classes) and CM-3 (3 classes). A mapping of segmentation labels is shown in Fig.~\ref{fig:class-mappings}.  

\section{Results}
\label{sec:result}

\begin{figure*}[htbp]
    \includegraphics[width=\linewidth]{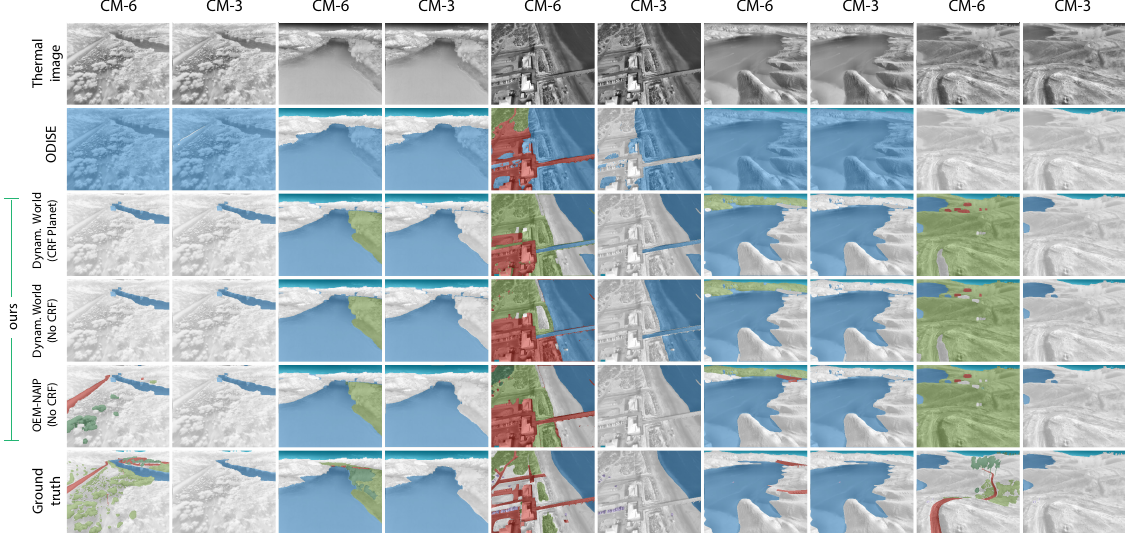}
    \caption{Generated segmentations from the baseline (ODISE~\cite{xu2023open}), our methods, and the ground truth (GT) using class mappings and colors from Fig. \ref{fig:class-mappings}. Mismatches between CM-6 labels and GT can occur depending on the LULC source used but are resolved with CM-3. Segmentations for classes containing small, sparse, and thin instances (CM-6), e.g. \textit{low vegetation} and \textit{built}, are hard to render due to low LULC resolution and low thermal contrast during label refinement.}
    \label{fig:results}
\end{figure*}

\subsection{Experimental Setup}
\subsubsection{Raster Acquisition}
We acquired \qty{10}{m} resolution Dynamic World LULC, 3D terrain data (\qty{3}{m} DEM, \qty{1}{m} DEM, \qty{2}{m} DSM) from USGS 3DEP, and high-resolution nadir imagery from NAIP (\qty{1}{m}) and Planet (\qty{3}{m}). Data was obtained via Microsoft Planetary Computer and Google Earth Engine.

\subsubsection{LULC from High-Resolution Imagery}
\label{sec:lulc-network-gen}
We used networks trained on Chesapeake Bay Program (CBP) and OpenEarthMap (OEM) datasets to produce two more high-resolution LULC sources alongside Dynamic World. For OEM-derived LULC, we used the pretrained U-Net model from~\cite{xia2023openearthmap}. To produce CBP-derived LULC, we fine-tuned a geospatial foundation model~\cite{mendieta2023towards} on the CBP dataset, using the 7-class set from~\cite{robinson2019large}.

We trained for \num{1000} epochs with a batch size of 16, a $1\mathrm{e}^{-3}$ learning rate, and RGB-NIR inputs of size 512$\times$512. 
To perform inference on large raster images, we use tiles with \qty{50}{\%} overlap and applied flips for test-time augmentation.

\subsubsection{LULC Refinement with Dense CRFs}
\begin{table}[tbp]
    \setlength{\aboverulesep}{1pt}
    \setlength{\belowrulesep}{1pt}
    \centering
    \caption{Evaluation of dense CRF refinement of Dynamic World LULC on NAIP imagery with ground truth labels from Chesapeake Bay Program (see class mapping in Fig.~\ref{fig:class-mappings}).}
    \label{tab:dense-crf-refinement}
    \resizebox{\linewidth}{!}{
    \begin{tabular}{lcccccccccc}
        \toprule[1pt]
         \multirow{2}{*}[-3pt]{\makecell[l]{CRF cond.\\source}} & \multirow{2}{*}[-3pt]{\makecell[c]{Boundary \\Loss~\textcolor{green}{$\downarrow$}}} & \multirow{2}{*}[-3pt]{mIoU~\textcolor{green}{$\uparrow$}} & \multicolumn{8}{c}{Dense CRF Parameters} \\ \cmidrule(lr){4-11}
         & & & $w_1$ & $w_2$ & $\theta_\gamma$ & $\theta_\alpha$ & $\theta_\beta^{\{0\}}$ & $\theta_\beta^{\{1\}}$ & $\theta_\beta^{\{2\}}$ & $\theta_\beta^{\{3\}}$  \\
         \midrule[1pt]
         None & 0.945 & 0.432 & --- & --- & --- & --- & --- & --- & --- & --- \\
         RGB$^\dagger$ & 0.914 & 0.441 & 1.00 & 1.00 & 200 & 195 & 7.00 & 7.00 & 7.00 & --- \\
         RGB & 0.777 & 0.452 & 47.2 & 2.63 & 33.3 & 149 & 1.14 & 1.14 & 1.14 & --- \\ 
         RGB-NIR & \textbf{0.749} & \textbf{0.453} & 47.4 & 0.14 & 61.5 & 194 & 128 & 0.22 & 125 & 2.71 \\ 
         \bottomrule[1pt]
    \end{tabular}
    }
    \vspace{-0.75mm}
    \caption*{\footnotesize$^\dagger$tuned by minimizing weighted cross entropy instead of boundary loss}
\end{table}
We refined the \qty{10}{m} Dynamic World LULC rasters on RGB-NIR imagery from NAIP and Planet (Fig.~\ref{fig:lulc-refinement}) using parameters from Tab.~\ref{tab:dense-crf-refinement}. Parameters were found using Bayesian optimization with Optuna~\cite{optuna_2019}. The search was done using NAIP as conditioning imagery and \qty{1}{m} resolution labels from CBP as ground truth (see Fig.~\ref{fig:class-mappings} for class mapping). For this use case, boundary loss was superior to standard cross-entropy loss (Tab.~\ref{tab:dense-crf-refinement}).

\subsubsection{Rendered Label Refinement}
For SAM refinement of the projected LULC labels, we used the default ViT-H model. We prompted with 32$\times$32 grid points and lowered the box non-maximum suppression threshold to 0.5. 

\subsubsection{Thermal Image Preprocessing}
We rescaled raw 16-bit thermal pixel intensities to sit between the 2\textsuperscript{nd} and 98\textsuperscript{th} percentiles before applying a contrast-limited adaptive histogram equalization with a 0.02 clip limit, following~\cite{lee2023online}. This was done for both visualization and algorithm input.

\begin{table*}[htp]
    \begin{minipage}{0.65\linewidth}
        \caption{LULC-generated semantic segmentation label assessment (mIoU) when compared to ground truth annotations, with comparisons against zero-shot visual foundation model baselines.}
        \label{tab:main-results}
        \setlength{\aboverulesep}{1pt}
        \setlength{\belowrulesep}{1pt}
        \resizebox{\linewidth}{!}{
        \begin{tabular}{lcc ccc  ccc}
            \toprule[1pt]
            \multirow{2}{*}[-2pt]{\makecell[l]{Method / \\LULC source}} & 
            \multirow{2}{*}[-2pt]{\makecell[c]{Dense CRF \\ refinement src.}} & 
            \multirow{2}{*}[-2pt]{\makecell[c]{3D source}} & \multicolumn{3}{c}{Dataset mIoU~\textcolor{green}{$\uparrow$}} & \multicolumn{3}{c}{Trajectory avg. mIoU~\textcolor{green}{$\uparrow$}} \\
            \cmidrule(lr){4-6} \cmidrule(lr){7-9}
            & & & \makecell[c]{CM-6} & \makecell[c]{CM-5} & \makecell[c]{CM-3} & \makecell[c]{CM-6} & \makecell[c]{CM-5} & \makecell[c]{CM-3}   \\
            \midrule[1pt]
            ODISE~\cite{xu2023open} & --- & --- & 0.299 & 0.262 & 0.330 & 0.264 & 0.304 & 0.413 \\
            OV-Seg~\cite{liang2023open} & --- & --- & 0.201 & 0.240 & 0.385 & 0.183 & 0.233 & 0.390 \\
            
            Chesapeake Bay (NAIP) & --- & DEM (3m) & 0.453 & 0.481 & 0.857 & 0.417 & 0.478 & 0.848 \\
            Chesapeake Bay (Planet) & --- & DEM (3m) & 0.236 & 0.305 & 0.657 & 0.201 & 0.251 & 0.555 \\
            Open Earth Map (NAIP) & --- & DEM (3m) & 0.549 & 0.562 & 0.868 & 0.440 & \textbf{0.528} & 0.864 \\
            Open Earth Map (Planet) & --- & DEM (3m) & 0.502 & 0.509 & 0.825 & 0.360 & 0.428 & 0.816 \\
            Dynamic World & --- & DEM (3m) & \textbf{0.577} & \textbf{0.572} & 0.876 & 0.450 & 0.518 & 0.860 \\
            
            Dynamic World & NAIP & DEM (3m) & 0.556 & 0.535 & 0.868 & 0.441 & 0.504 & 0.865 \\
            Dynamic World & Planet & DEM (3m) & 0.573 & 0.557 & \textbf{0.887} & \textbf{0.455} & 0.510 & \textbf{0.870} \\
            \bottomrule[1pt]
        \end{tabular}
        }
    \end{minipage}
    \hfill
    \begin{minipage}{0.335\linewidth}
        \setlength{\aboverulesep}{1pt}
        \setlength{\belowrulesep}{1pt}
        \caption{Ablation studies}
        \label{tab:ablation}
        \begin{subtable}{\linewidth}
            \vspace{-1mm}
            \caption{3D source ablation}
            \vspace{-1mm}
            \label{tab:3d-source-ablation}
            \resizebox{\linewidth}{!}{
            \begin{tabular}{lc ccc}
                \toprule[1pt]
                \multirow{2}{*}[-2pt]{\makecell[l]{Method}} & 
                \multirow{2}{*}[-2pt]{\makecell[c]{3D source}} & 
                \multicolumn{3}{c}{Traj. avg. mIoU} \\
                \cmidrule(lr){3-5}
                & & \makecell[c]{CM-6} & \makecell[c]{CM-5} & \makecell[c]{CM-3} \\
                \midrule[1pt]
                \multirow{3}{*}{\makecell[l]{Dynamic World \\ + SAM}} & DEM (3m) & \textbf{0.450} & \textbf{0.518} & \textbf{0.860} \\
                 & DEM (1m) & 0.441 & 0.507 & 0.842 \\
                 & DSM (2m) & 0.439 & 0.504 & 0.848 \\
                \bottomrule[1pt]
            \end{tabular}
            }
        \end{subtable}

        \vspace{1.5mm}
    
        \begin{subtable}[b]{\linewidth}
            \caption{Label refinement ablation}
            \vspace{-1mm}
            \label{tab:label-refinement-ablation}
            \setlength{\aboverulesep}{1pt}
            \setlength{\belowrulesep}{1pt}
            \resizebox{\linewidth}{!}{
            \begin{tabular}{lc ccc}
                \toprule[1pt]
                \multirow{2}{*}[-2pt]{\makecell[l]{Method}} & 
                \multirow{2}{*}[-2pt]{\makecell[c]{Projected label \\ refine method}} & 
                \multicolumn{3}{c}{Traj. avg. mIoU} \\
                \cmidrule(lr){3-5}
                & & \makecell[c]{CM-6} & \makecell[c]{CM-5} & \makecell[c]{CM-3} \\
                \midrule[1pt]
                \multirow{3}{*}{\makecell[l]{Dynamic World \\ + DEM (3m)}} & SAM & \textbf{0.450} & \textbf{0.518} & \textbf{0.860} \\
                 & SLIC & 0.392 & 0.452 & 0.711 \\
                 & Felzenszwab & 0.369 & 0.426 & 0.677 \\
                \bottomrule[1pt]
            \end{tabular}
            }
        \end{subtable}
    \end{minipage}
    \vspace{-0.25cm}
\end{table*}

\subsection{Satellite-based Semantic Segmentation Label Generation}
\label{sec:main-results}
We compare our LULC-generated semantic segmentation labels to manually-annotated, ground truth labels. Due to class differences between LULC data and ground truth, we evaluate on three ground truth-derived class sets of increasing generality (CM-6, CM-5, CM-3). We report the overall dataset mIoU and the trajectory-averaged mIoU in Tab.~\ref{tab:main-results}.

Overall, our method delivers thermal semantic segmentation labels consistent with ground truth (Fig.~\ref{fig:results}). Notably, our best variants greatly outperform the zero-shot semantic segmentation models, ODISE~\cite{xu2023open}, and OV-Seg~\cite{liang2023open}, which were prompted with a list of classes present in the dataset. We note that ODISE and OV-Seg are occasionally effective on thermal images, but lack consistency.

Among our methods without LULC refinement, semantic segmentation label generation using Dynamic World and DEM (\qty{3}{m}) as a 3D source generally outperforms other variants using CBP- and OEM-derived LULC sources. LULC created from the OEM network on NAIP data provides improvements (0.005 - 0.01 mIoU) in trajectory-averaged mIoU over Dynamic World for the CM-5 and CM-3 class sets. Despite marginal gains, this is likely due to the higher resolution (\qty{1}{m}) of OEM/NAIP-derived LULC, which enables segmentation renderings of small or thin classes that are present in CM-5, such as roads (see Fig.~\ref{fig:results}). This is not possible with Dynamic World due to its lower \hbox{\qty{10}{m} resolution}. 

Conversely, LULC generated from Planet imagery provides poor results due to domain differences between OEM/CBP training images and Planet.
When used for dense CRF refinement of \qty{10}{m} Dynamic World rasters, however, Planet imagery uniquely provides $\sim$0.01 boost for both mIoU metrics on the most general CM-3 class set. This behavior is absent when refining on NAIP imagery due to terrain changes between thermal and NAIP acquisition dates.

\begin{figure}[htbp]
    \centering
    \includegraphics[width=\linewidth]{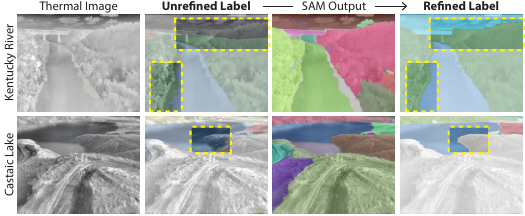}
    \caption{Rendered label refinement process with SAM~\cite{kirillov2023segany}.}
    \label{fig:unrefined_vs_refined}
\end{figure}

Furthermore, we note that our method can handle temporal mismatches between satellite and thermal data even as environments naturally evolve. For example, coastal tide patterns and varying lake levels (Fig. \ref{fig:unrefined_vs_refined}, Castaic Lake) may shift class boundaries within short time periods. Due to SAM's ability to capture entire class instances, our rendered label refinement step (Sec.~\ref{sec:renderedlabelrefinement}) is notably able to overcome such changes as long as most of the true class is still rendered.

Due to its accessibility and competitive performance, we advocate using Dynamic World LULC for satellite-based semantic segmentation label generation efforts, with potential refinement via temporally-relevant, high resolution imagery. However, this will inevitability change with advancements in LULC creation and as sub-meter data products with high temporal coverage become more freely-accessible.  

\subsection{Ablation Study}
In these ablations, we use Dynamic World as our semantic source. Unless otherwise specified, we use \qty{3}{m} DEMs to add 3D context and do not use any CRF refinement.

\subsubsection{3D Data Source}
First, we compare LULC-based semantic segmentation label generation with \qty{3}{m} DEMs against two additional 3D data sources: \qty{2}{m} DSMs and \qty{1}{m} DEMs. Due to limited coverage, we lack DSMs and \qty{1}{m} DEMs over the thermal data capture areas of Colorado River and Duck, respectively, and resort to \qty{3}{m} DEMs in those areas. Our results show that \qty{3}{m} DEMs provide consistently higher trajectory-averaged mIoU across all three class sets, despite the other two sources supposedly providing more accurate and precise 3D terrain data (Tab.~\ref{tab:3d-source-ablation}). Reasons for this may include temporal differences or spatial misalignment during orthorectification. Nonetheless, all 3D sources generally perform well and any one of these 3D products can be used for our method when the other two are unavailable.

\subsubsection{Raster Spatial Resolution}
\begin{figure}[ht]
    \centering
    \includegraphics[width=\linewidth]{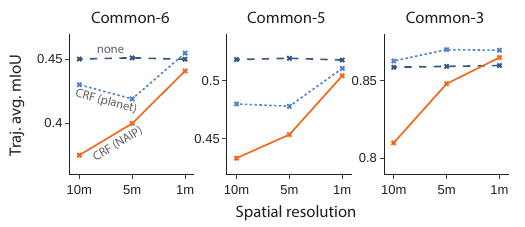}
    \caption{Effect of LULC spatial resolution on semantic segmentation label generation.}
    \label{fig:ablation-spatial-res}
\end{figure}
To assess the impact of LULC spatial resolution on label generation, we generate labels from Dynamic World LULC rasters resampled to \qty{10}{m} (native), \qty{5}{m}, and \qty{1}{m} resolution. We use nearest neighbor interpolation on the LULC directly, and CRF refinement on NAIP and Planet rasters (resampled to \qty{10}{m}, \qty{5}{m}, and \qty{1}{m} resolutions via bicubic interpolation). 

Our results (Fig.~\ref{fig:ablation-spatial-res}) suggest that LULC spatial resolution matters more for more specific class sets (CM-6/CM-5), and becomes less critical as class sets generalize \hbox{(CM-3)}. Moreover, we find greater benefits from CRFs when conditioning on higher-resolution imagery, especially when dealing with the larger and more specific class sets (CM-6/CM-5). This is likely due to smoothing over small or thin class instances that comprise of a few pixels when refining at lower resolutions.

\subsubsection{Segment Anything vs. Classical Methods for Projected LULC Refinement}
We compare our choice of SAM for projected LULC label refinement against SLIC~\cite{achanta2012slic} and Felzenszwab~\cite{felzenszwalb2004efficient} superpixels. We use implementations from \texttt{scikit-image}~\cite{scikitimage}, setting SLIC's number of segments to 100 and compactness to 10, and Felzenszwab's scale parameter to $1\mathrm{e}^4$. We select these parameters to maximize segmentation area while remaining within class instances. 

Overall, SAM consistently outperforms the other methods, with mIoU margins increasing from 0.06 (Tab.~\ref{tab:label-refinement-ablation}). This is because SAM can produce semantically distinct masks in the thermal domain, albeit less reliably than in RGB. This allows minor imperfections to be ignored through majority vote (Alg.~\ref{alg:sam-refinement}). In contrast, classical methods produce fragmented, semantic-agnostic masks which offer little benefit. 

\begin{figure}[htbp]
    \centering
    \includegraphics[width=\linewidth]{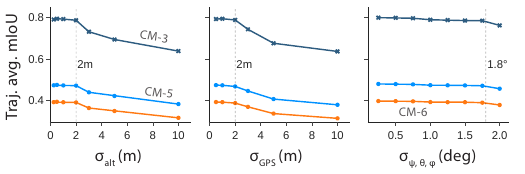}
    \caption{Effect of global pose estimate precision on semantic segmentation label generation with SAM refinement.}
    \label{fig:ablation-sensor-error}
\end{figure}
\subsubsection{State Estimate Precision}
To quantify the effect of global pose estimation precision on our label generation process, we systematically perturb these measurements by sampling from a normal distribution with increasing variance. Our analysis reveals that, with \qty{95}{\%} confidence, label generation remains robust for global positioning and altitude estimates within roughly \qty{4}{m} and for orientations within roughly 3.5\textdegree~(Fig.~\ref{fig:ablation-sensor-error}). These findings are consistent across class sets.
During development, both synchronizing the timing of image capture to the IMU data was shown to be critical, as was the SAM refinement stage for compensation for attitude estimate errors (see Kentucky River in Fig. \ref{fig:unrefined_vs_refined}). 

\subsection{Application: Semantic Segmentation Model Training}
To demonstrate our method for field robot perception, we trained an EfficientViT-B0 semantic segmentation network~\cite{cai2023efficientvit} using the aerial thermal dataset and general train/val/test split from~\cite{lee2024cart}. Three sets of labels (CM-6, CM-5, and CM-3) were generated for training and validation using our method, with ground truth labels converted accordingly for testing and baseline training. All networks were trained following the thermal training procedure from~\cite{lee2024cart}.

Our semantic segmentation results (Tab.~\ref{tab:application-all}) closely match the mIoU of the generated annotations (Tab.~\ref{tab:main-results}). Networks trained with CM-3 classes resulted in 0.889 mIoU during testing, compared to 0.962 mIoU for those trained with ground truth labels. Networks trained on CM-5 and \hbox{CM-6} show larger gaps (Tab.~\ref{tab:application-all}) but still show the benefit of our method. We find this is largely due to difficulties in accurately rendering land-based classes, specifically \textit{low vegetation} and \textit{built} (Tab.~\ref{tab:application-cm6}). These classes contain small and thin entities like sparse shrubs or roads, and are not always precisely shown in LULC data. Also, they can be missed during rendered label refinement (Sec.~\ref{sec:renderedlabelrefinement}) due to blurred and low-contrast appearance in thermal imagery. Despite this, our method can effectively train semantic segmentation models, particularly with the CM-3 class set, and support field robotic applications like nighttime river navigation~\cite{lee2023online}.

\begin{table}[htbp]
\setlength{\aboverulesep}{1pt}
\setlength{\belowrulesep}{1pt}
    \centering
    \caption{Test results (mIoU) of semantic segmentation networks trained on LULC-generated labels and networks trained on manually-annotated ground truth.}
    \label{tab:application}
    \begin{subtable}{\linewidth}
        \centering
        \caption{Segmentation results after training on CM-6 (least inclusive), CM-5, and CM-3 (most inclusive) class sets.}
        \label{tab:application-all}
        \resizebox{\linewidth}{!}{
        \begin{tabular}{p{3cm}P{1.35cm}P{1.35cm}P{1.35cm}}
        \toprule[1pt]
        \multirow{2}{*}[-2pt]{Annotation Method} & \multicolumn{3}{c}{Class set} \\
        \cmidrule(lr){2-4}
        & CM-6 & CM-5 & CM-3 \\
        \midrule[1pt]
        LULC-generated & 0.542 & 0.547 & 0.889 \\
        Ground truth & 0.819 & 0.836 & 0.962 \\
        \bottomrule[1pt]
        \end{tabular}
        }
    \end{subtable}
    
    \vspace{2mm}
    
    \begin{subtable}{\linewidth}
        \caption{Per-class IoU for networks trained using the CM-6 class set.}
        \label{tab:application-cm6}
        \resizebox{\linewidth}{!}{
        \begin{tabular}{p{2cm}cccccc}
        \toprule[1pt]
        Annot. Method & water & trees & low veg. & built & ground & sky \\
        \midrule[1pt]
        Generated & 0.880 & 0.529 & 0.165 & 0.289 & 0.521 & 0.868 \\
        Ground truth & 0.963 & 0.787 & 0.702 & 0.653 & 0.854 & 0.955 \\
        \bottomrule[1pt]
        \end{tabular}
        }
    \end{subtable}
    
\end{table}

\subsection{Computational Costs and Pricing}
Our method annotates a single image in \qty{3}{s}, \qty{2.86}{s} of which is due to SAM. Annotations are \textit{free} when using only Dynamic World LULC but cost $\sim$\$10 USD/$\mathrm{km}^2$ with CRF refinement due to the price of realtime, high-resolution satellite imagery. With our method, annotating 2\,000 images takes 1.6 hours on a single workstation, in contrast with the usual 2-4 week timeframe and \$3\,000 to \$8\,000 USD outsourcing cost\footnote{\$1.50 to \$4.00 per image for 1-10 semantic segmentation classes, based on pricing from Scale AI at the time of writing: \href{https://scale.com/pricing}{https://scale.com/pricing}}. We note that CRF refinement can be cost-effective for large data volumes in a concentrated area due to its one-time cost, but \qty{98.5}{\%} of its performance (CM-6, CM-3) is achievable with free \qty{10}{m} resolution LULC (Sec.~\ref{sec:main-results}).

\section{Conclusion}
\label{sec:conclusion}
We presented a novel method for automatically generating high-quality semantic segmentation annotations for classes often encountered by aerial robots in field settings. Our approach leverages satellite data products and employs refinement steps to achieve fine precision at class boundaries even with low-resolution satellite data, achieving 98.5\% of the performance of costly high-resolution options. We demonstrated the robustness of our method to global positioning and attitude estimation errors, indicating that it can provide good segmentations even with inexpensive sensors and slight data desynchronization, and identified limitations due to small and thin class instances. Lastly, we demonstrated its application to field robot perception by successfully training a semantic segmentation network solely with generated labels. 
This method enables rapid training of thermal perception stacks using incremental learning as new field data is collected.

\end{document}